\documentclass{ifacconf}
\usepackage{graphicx}
\usepackage{natbib} 
\usepackage{amsmath,amssymb,amsfonts,mathtools}
\usepackage{xcolor}

\DeclareMathOperator*{\argmax}{arg\,max}

\DeclareMathOperator{\relu}{ReLU}

\usepackage{tikz}                   
\usepackage{tikzscale}              
\usetikzlibrary{calc}               
\usetikzlibrary{patterns,shapes} 

\newtheorem{theorem}{Theorem}

\begin{document}
\begin{frontmatter}

\title{An Overview and Prospective Outlook on Robust Training and Certification of Machine Learning Models}

\thanks[footnoteinfo]{This work was supported by grants from ONR and NSF.}

\author[berkeley,equal]{Brendon G.\ Anderson} 
\author[berkeley,equal]{Tanmay Gautam} 
\author[berkeley]{Somayeh Sojoudi}

\address[berkeley]{University of California, Berkeley, USA (e-mails: \{bganderson,tgautam23,sojoudi\}@berkeley.edu)}
\address[equal]{Equal contribution}

\begin{abstract} 
In this discussion paper, we survey recent research surrounding robustness of machine learning models. As learning algorithms become increasingly more popular in data-driven control systems, their robustness to data uncertainty must be ensured in order to maintain reliable safety-critical operations. We begin by reviewing common formalisms for such robustness, and then move on to discuss popular and state-of-the-art techniques for training robust machine learning models as well as methods for provably certifying such robustness. From this unification of robust machine learning, we identify and discuss pressing directions for future research in the area.
\end{abstract}

\begin{keyword}
Neural networks, learning theory, robustness analysis, large-scale optimization, relaxations
\end{keyword}

\end{frontmatter}

\section{Introduction}
Over the past decades, the rise in computing power and available data has propelled the field of machine learning research. Nowadays, machine learning technology is ubiquitous with vast applications ranging from image classification to speech recognition to autonomous driving \citep{madry2018towards,Goodfellow-et-al-2016}.

The successful deployment of machine learning systems in real-world control systems requires both safety and reliability guarantees on the underlying models. This in turn necessitates that the learned models be robust towards processing corrupted data at both training and testing time. These corruptions in the data can typically manifest as either nature-driven or meticulously-crafted, malicious perturbations and are often modeled as additive noise \citep{biggio2013evasion,szegedy2014intriguing}. In either case, an effective means of ensuring reliability of the model is by factoring in the \textit{worst-case} noise perturbations during the training procedure.

To illustrate the vitalness of robustness against noise perturbations, we look towards the field of computer vision. Here studies have shown that slight manipulations in the input images can elicit misclassifications in neural networks with high confidence \citep{szegedy2014intriguing, deepfool,goodfellow2015explaining}. Thus, robustness of the underlying models is crucial to safety-critical technologies such as autonomous driving, where models susceptible to subtle manipulations in the data could result in grave consequences \citep{kurakin2017adversarial}.

In this discussion paper, we overview the problems surrounding robustness of machine learning models as well as recently proposed solutions in the literature, both from the perspective of training and testing, and we identify directions for future research at this front. Specifically, we introduce the robust training framework in Section \ref{sec: robust_training_framework}, discuss adversarial attacks and the related adversarial training approach in Section \ref{s:art}, and present methods for provably certifying the robustness of trained models in Section \ref{sec: certification}. We conclude and give prospective outlooks on the research area in Section \ref{sec: conclude}.

\subsection{Notations}
The real numbers are denoted by $\mathbf{R}$, and we use the shorthand notation $[N]$ to mean $\{1,2,\dots,N\}$. If $x$ is a random variable with probability distribution $\mathcal{X}$ and $f$ is a measurable function, we denote the expectation of $f(x)$ by $\mathbf{E}_{x\sim\mathcal{X}}f(x)$. The normal distribution with mean $\overline{x}$ and covariance matrix $\Sigma$ is denoted by $N(\overline{x},\Sigma)$. We use $\|\cdot\|$ to denote an arbitrary norm on $\mathbf{R}^d$, whereas the $\ell_p$-norms are defined by $\|x\|_p = \left(x_1^p + \cdots + x_d^p \right)^{1/p}$ for $p\in[1,\infty)$ and $\|x\|_\infty = \max\{|x_1|,\dots,|x_d|\}$. Recall that the rectified linear unit ($\relu$) is defined by $\relu(x) = \max\{0,x\}$, which is taken element-wise for vectors.
\section{Robust Training Framework}
\label{sec: robust_training_framework}
As machine learning models become increasingly prevalent in real-world systems, it is insufficient for the systems to function as expected for \textit{most} of the time---they should be truly reliable and robust at \textit{all} times accounting for worst-case situations. In this section, we first introduce the general frameworks of robust optimization and supervised learning. Subsequently we outline several notions of robustness relevant to modern machine learning systems. 

\subsection{Robust Optimization}
The basic idea towards ensuring the robustness of machine learning models is to take the inherent uncertainty in the input data into consideration during the training or learning process. This process is known as \textit{robust training} or \textit{robust learning} and it draws upon concepts from the more traditional field of robust optimization \citep{soyster, Ghaoui1997RobustST, bertsimasprice, 2000rolinear}.

We consider a general optimization problem of the form
\begin{equation}
\begin{aligned}
    & \underset{x}{\text{minimize}} && f(x) \\
    & \text{subject to} && g_i(x)\leq 0, ~ i\in\{1,2,\dots,n\}, \\
    &&& h_j(x)=0, ~ j\in\{1,2,\dots,m\},
    \end{aligned}
    \label{eq:generaloptimization}
\end{equation}
where $f$ is the objective function, $g_i$'s are the inequality constraint functions, and $h_j$'s are the equality constraint functions. In order to incorporate data uncertainty into optimization (\ref{eq:generaloptimization}), robust optimization assumes that the objective and constraint functions lie in sets within function space called \textit{uncertainty sets}. The goal of robust optimization is obtain a solution that is feasible to all constraint function perturbations possible within the uncertainty sets, while ensuring optimality for the worst-case objective function.   

Due to the inherent uncertainty in the real-world and coupled with advances of convex optimization solvers, the field of robust optimization has found many applications in various domains including control, finance, operations research and machine learning \citep{boyd2004convex, bertsimasRO}. Robust optimization has been utilized in the context of stability analysis/synthesis when dealing with uncertain dynamic systems \citep{ro_method_applications}.  Within finance, robust optimization lends itself naturally to the problem of portfolio optimization, where asset returns have an associated risk \citep{BEN:09}. Studies have also linked the notion of regularization in machine learning with robust optimization. For example, the work \citet{Ghaoui1997RobustST} shows that the solution to regularized least squares regression coincides with the solution of a robust optimization problem. \cite{robustsvm} show a similar result for regularized support vector machines (SVM).

Before we move on, we would also like to briefly introduce the closely aligned field of stochastic optimization (SO) and clarify differences with robust optimization. In SO, the uncertain problem data is assumed to be random with the underlying distributions known \textit{a priori} or in more advanced settings partially known \citep{BEN:09}. Therefore, SO distinguishes itself from robust optimization as the latter assumes a deterministic and set-based uncertainty model \citep{bertsimasRO}. The SO setting is generally known to be less conservative than the worst-case setting enforced by robust optimization.

\subsection{Supervised Learning}
Supervised learning is the prominent framework within machine learning to obtain predictors from a given dataset. Consider a stationary distribution of samples $\mathcal{X}$ and labels $\mathcal{Y}$ with a true predictor $f^\star$ that maps a particular sample to a corresponding label, i.e., $f^\star:\mathcal{X}\to \mathcal{Y}$. As an example, for the task of image classification, $\mathcal{X}$ can be the support of the distribution of all images and $\mathcal{Y}$ can be the associated image categories. Furthermore, suppose we have independently drawn $N$ sample-label pairs $\{(x_i, y_i)\}_{i=1}^N$ where $x_i\sim \mathcal{X}, y_i\sim\mathcal{Y}$ for $i\in [N]$. These data samples are \textit{independent and identically distributed}. The goal of supervised learning is to find the true mapping $f^\star$ of samples to labels in the absence of knowledge of the distributions $\mathcal{X}$ and $\mathcal{Y}$, while using only the drawn dataset. The task of find a good predictor under these constraints can be cast as an optimization problem known as empirical risk minimization (ERM). Fundamental to ERM is the notion of a loss function $\ell:\mathcal{Y}\times \mathcal{Y}\to\mathbf{R}$ that measures the discrepancy between the output of the assumed predictor and the true label. To pose a tractable optimization problem, the search of a good predictor must be restricted to a function class $\mathcal{F}\subseteq \{f\colon \mathcal{X}\to\mathcal{Y}\}$. The empirical risk minimization problem is given as follows:
\begin{equation}\label{eq:erm}
    \inf_{f\in\mathcal{F}} \sum_{i=1}^N\ell (f(x_i), y_i).
\end{equation}

Note that ERM minimizes the loss taking into account the training data as is. There is no notion to consider potential variations or perturbations in the data at hand. This suggests a lack of reliability in the learned model and requires that we couple the ERM problem with concepts from robust optimization to yield more robust models. In the next subsections, we outline ways that uncertainty in the input data can manifest in the context of modern day machine learning.

\subsubsection{Distribution Shift.}
The supervised learning paradigm relies on the key assumption that the training and test data are drawn from the same distribution. In many practical applications, however, this is not the case. Examples of this include performative prediction \citep{pmlr-v119-perdomo20a} and behavioral cloning \citep{dagger} in reinforcement learning. Performative prediction studies the phenomenon where prediction models are used as a means to support real-world decisions \citep{pmlr-v119-perdomo20a}. However, when these decisions manifest, they may change the dynamics of the underlying environment such that the prediction model is no longer relevant. In behavioural cloning, a decision-making policy is learned using samples generated by an expert. However, as the learned policy is merely an approximation of the expert policy, the underlying policy model has to tackle the challenge of making decisions on out-of-distribution samples \citep{dagger}.

More generally, machine learning models are trained with a training data batch that best represents the deployment environment. However, the considered training dataset may not fully capture the entire complexity of the real-world environment at hand, and therefore at deployment the model is tasked with inputs drawn from a different distribution as compared to training time. As an illustrative example, consider a complex real-world environment and a simplistic simulation thereof. When real world data is too expensive to collect, such simulations offer a convenient way of gathering training data. However, since the simulation is only an approximation of the real world and the latter is subject to temporal changes, the underlying distributions that generate the simulated and real-world data differ from one another. Machine learning models need to be robust to such distribution shifts. 

In distributionally robust optimization (DRO) \citep{scarf, datadrivenro}, instead of choosing a model $f$ that minimizes the empirical risk (\ref{eq:erm}), it is assumed that an adversary can perturb the underlying data distribution $\mathcal{X}$ within a set $\mathcal{U}$ centered around the empirical distribution $\hat{\mathcal{X}}_N = \frac{1}{N}\sum_{i=1}^N \delta_{x_i}$ where $\delta_{x_i}$ represents a discrete unit weight centered at $x_i$. DRO tries to find a performative model that accounts for all perturbations of $\mathcal{X}$:
\begin{equation}\label{eq:dro}
    \inf_{f\in\mathcal{F}}\sup_{\tilde{\mathcal{X}}\in \mathcal{U}} \mathbf{E}_{x\sim \tilde{\mathcal{X}}} \ell(f(x), y).
\end{equation}
The robustness obtained from (\ref{eq:dro}) can directly yield improved generalization: if the data that forms the empirical distribution $\hat{\mathcal{X}}_N$ is drawn from a population distribution $\mathcal{X}$ and $\mathcal{X}\in\mathcal{U}$, then the optimization (\ref{eq:dro}) accounts for the true distribution and upper bounds the out of sample performance \citep{NEURIPS2019_1770ae9e}. In practice, $\mathcal{U}$ is selected as a ball of radius $\epsilon$ away from the empirical distribution $\hat{\mathcal{X}}_N$; $\mathcal{U} = \{\tilde{\mathcal{X}}: d(\tilde{\mathcal{X}},\hat{\mathcal{X}}_N)\leq \epsilon\}$. Here, $d(\cdot, \cdot)$ is a measure of divergence between two distributions. Careful selection of $d(\cdot, \cdot)$ is vital, as it greatly impacts the tractability of the DRO problem. Within the context of machine learning, the main divergence measures considered are $\phi$-divergences \citep{dro1,dro2} and the Wasserstein distance \citep{dro3, dro4,dro5}.

\subsubsection{Adversarial Robustness.}
While many recent works of the last few years have studied the notion of distributional robustness, it is in fact adversarial examples that initially put the limelight on robustness in modern deep learning. Recent works have categorized the relationship between these two notions of robustness. In \citep{NEURIPS2019_1770ae9e,sinha2018certifiable,Staib2017DistributionallyRD}, the authors suggest that distributional robustness can be seen as a generalization to adversarial robustness.

Studies have revealed how carefully-crafted adversarial examples can exploit deficiencies in modern machine learning systems and elicit undesired outputs. This is especially true in the context of state-of-the-art deep learning classifiers \citep{szegedy2014intriguing, goodfellow2015explaining, madry2018towards}. As deep learning has emerged as the predominant subfield within machine learning with a vast array of applications, we devote the entirety of Section \ref{s:art} to adversarially robust training in the context of this family of models. 
\section{Adversarially Robust Training}\label{s:art}
In this section, we introduce the framework of adversarially robust training, which primarily addresses the challenge of dealing with corrupt data at test time. We begin the section with an overview of what adversarial examples are, highlighting their properties and why deep learning architectures are susceptible. This is followed by a comprehensive overview of both adversarial attack and adversarial training methods.

\subsection{Adversarial Examples}
\subsubsection{Definition.}\label{sss:definition} 
\begin{figure}[htbp]
\centering
\includegraphics[width=0.9\linewidth]{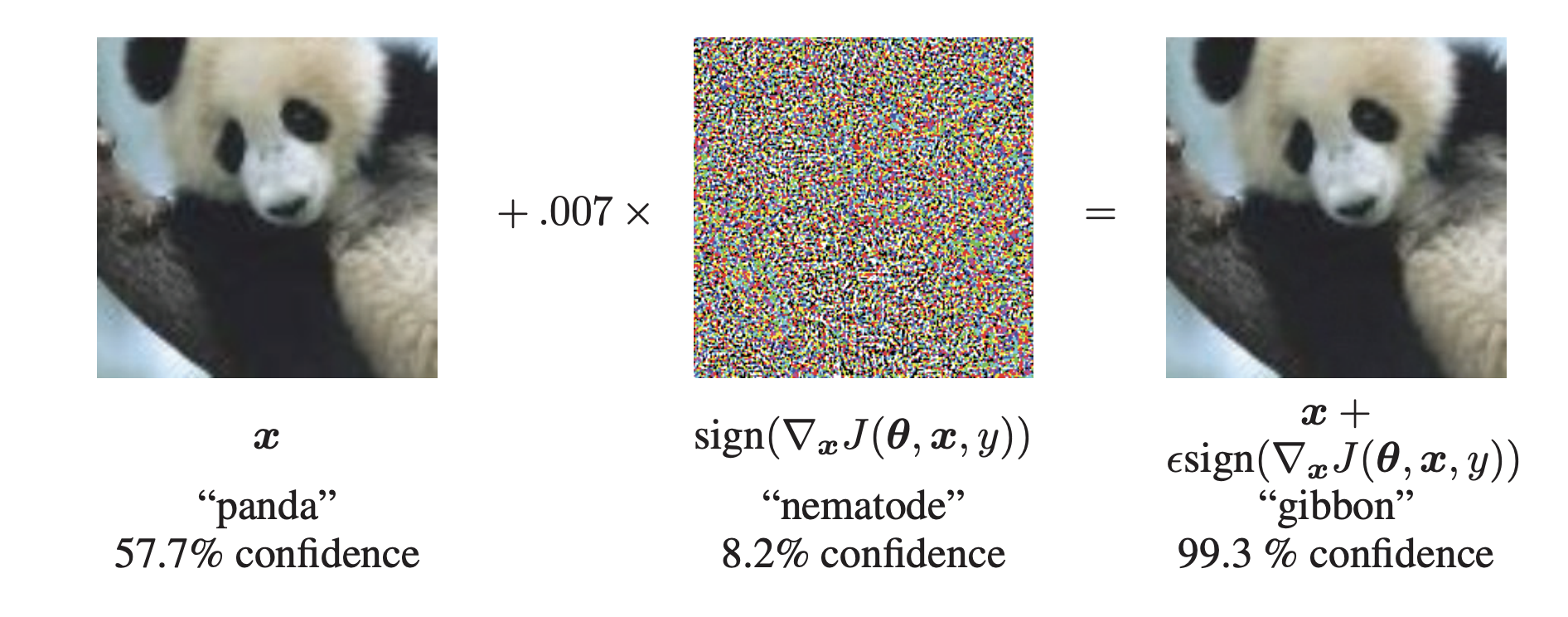}
\caption{Example of an imperceptibly  perturbed adversarial example \citep{goodfellow2015explaining}}
\label{fig:imperceptibleexample}
\end{figure}

\begin{figure}[htbp]
\centering
\includegraphics[width=0.6\linewidth]{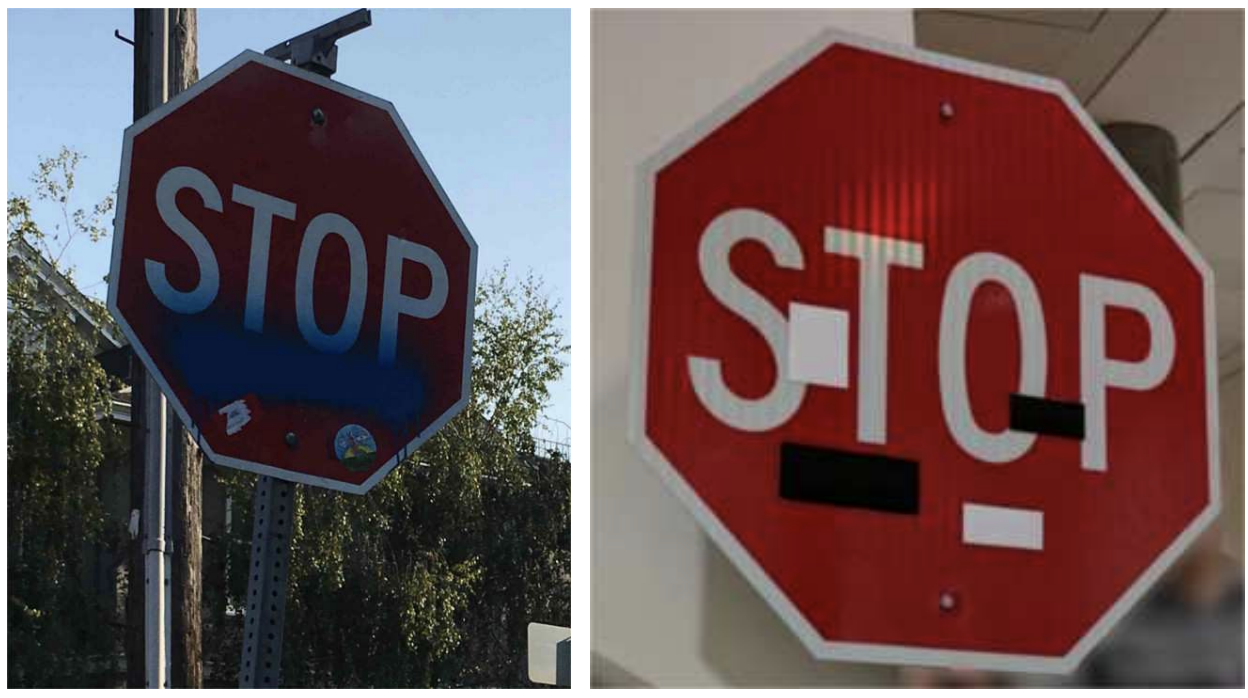}
\caption{Example of an inconspicuously perturbed adversarial example \citep{Evtimov}}
\label{fig:inconspicuousexample}
\end{figure}

Informally, an adversarial example is an ``imperceptibly'' or ``inconspicuously'' perturbed sample that would elicit misclassification from the considered model \citep{Wiyatno2019AdversarialEI}. More formally, consider a nominal sample $x$ and a slightly perturbed sample $x^\prime$. We define $x^\prime$ to be an adversarial example when
\begin{equation}
\text{$d(x, x^\prime)\leq \epsilon$ and $f(x)\neq f(x^\prime)$,}
\label{eq:adversarial_example_definition}
\end{equation}
where $d(\cdot, \cdot)$ is some distance metric, $\epsilon >0$ is some small positive scalar, and $f$ is the considered model. Here, $d(\cdot, \cdot)$ can be some standard distance metric on $\mathbf{R}^d$, e.g., the $\ell_p$-norm. While the usual treatment of adversarial examples is that these are subject to imperceptible modifications (i.e., $\epsilon$ is very small), they have also been studied with regards to visible but non-obvious inconspicuous changes \citep{46561,Evtimov,Sharif2016AccessorizeTA}. In the latter, the constraint upon $d(x,x')$ is slightly relaxed. An example of an imperceptibly perturbed image is shown in Figure \ref{fig:imperceptibleexample}. Here, imperceptible noise is added to an image of a panda, which is subsequently misclassified as a gibbon. On the other hand, an inconspicuously perturbed image is depicted in Figure \ref{fig:inconspicuousexample}. Despite the fact that the stop sign is partially occluded by stickers, it doesn't diminish a human's ability to recognize the stop sign. On the other hand, deep learning models have been shown to be susceptible to such visible perturbations \citep{Evtimov}.    

\subsubsection{Properties.}
\begin{enumerate}
    \item Imperceptibility: As described above, adversarial examples are typically designed to be such that their difference to the original sample is imperceptible to humans. As previously highlighted, this property can be relaxed to yield inconspicuous adversarial examples, where there is a visible change that does not alter the semantic features of the sample.
    \item Cross-Model Transferability: Adversarial examples exhibit the property of cross-model transferability, whereby they can mislead different deep learning models trained on the same dataset \citep{szegedy2014intriguing,goodfellow2015explaining,transferability}.
    \item Cross Training-Set Transferability: Adversarial examples can also be carefully designed to fool classifiers without directly modifying training samples. Models cannot guard against such examples by augmenting perturbed training samples.
\end{enumerate}

Now that we have established what adversarial examples are, the question as to why deep learning models are particularly susceptible to these still remains. While this is still an ongoing research theme, studies have outlined several possible reasons for this vulnerability. In \citet{Arpit}, the authors measure the memorization capacity of deep learning models and show that models with a greater degree of memorization are increasingly vulnerable to adversarial examples. In \citet{Jo2017MeasuringTT}, authors verify that convolutional architectures learn the surface statistical regularities of the training dataset, rather than abstract high-level representations. This provides a possible explanation for why adversarial examples may mislead the networks. Moveover, this study gives insight into the transferability property of adversarial examples: different models trained on the same dataset learn the same statistical regularities leading to them being fooled by the same adversarial example. Another line of reasoning as to why deep learning models are fallible is suggested in \citet{Adversarialfeatures}, where the authors claim that models learn non-robust features during the training process.

\subsection{Adversarial Attacks}
\subsubsection{Dichotomy of Adversarial Attacks.}
Adversarial attacks are mainly categorized as white-box and black-box attacks. In the former, the attacker has full knowledge of the target model, e.g., with regards to its architecture, parameters, and gradient information. On the contrary, black-box attacks only have knowledge of the output of the target model. While in practice the black-box assumption is more realistic due to attackers not usually having full information on the target, white-box attacks are vital to consider to guard against the worst-case scenario.

Adversarial attacks can be alternatively characterized as targeted and non-targeted attacks. Here, targeting refers to specific manipulation of the output of the model. In targeted attacks, the attacker aims to design an adversarial example in a manner which causes the model to output a specific classification. Non-targeted attacks aim to elicit any form of misclassification.

\subsubsection{White-Box Optimization-Based Attacks.}\label{sss:whiteboxoptimization}
We first consider white-box optimization-based attacks, where the idea is to formulate an optimization problem involving the model's loss function to directly find an adversary.

The Limited Memory Broyden-Fletcher-Goldfarb-Shanno (L-BFGS) \citep{szegedy2014intriguing} attack, is one of the earliest adversarial attack methods which was able to fool the then state-of-the-art AlexNet \citep{alexnet} and QuocNet architectures on image classification tasks. In this method of attack, the goal is to solve the optimization
\begin{equation}\label{eq:lbfgs_attack1}
    \min_{x^\prime : f(x^\prime)\neq f(x)} \|x-x^\prime\|_2,
\end{equation}
where as before $x$ is the non-adversarial sample and $f$ is the model. Solving this optimization essentially yields the sample $x^\prime$ of smallest perturbation that elicits misclassification. Due to the difficulty in solving the optimization \eqref{eq:lbfgs_attack1}, the authors of \citet{szegedy2014intriguing} reformulated the problem into a box-constrained one:
\begin{equation}\label{eq:lbfgs_attack2}
    \min_{x^\prime\in [0,1]^d} c\|x-x^\prime\|_2 + \mathcal{L}(x^\prime, t),
\end{equation}
where $c>0$ is a scalar, $\mathcal{L}(\cdot, \cdot)$ represents the loss function of the target model (e.g., cross-entropy loss), $t$ is the target misclassification label, and the non-adversarial sample $x$ has been normalized to lie within the cube $[0,1]^d$. Problem \eqref{eq:lbfgs_attack2} can be solved with the 2nd-order L-BFGS optimization algorithm giving the attack method its name. As the problem doesn't have an explicit misclassification constraint, the solution is not guaranteed to be an adversarial example. The constant $c$ can be iteratively increased via line search until an adversary is obtained.

The L-BFGS attack \citep{szegedy2014intriguing} generates adversarial examples by optimizing over the loss between the model output and target class. An alternative formulation is to optimize over the similarity between the hidden representations of the perturbed sample and a target sample with a different class. The adversarial manipulation of deep representations (AMDR) method \citep{Sabour2016AdversarialMO} attempts to do exactly this. More formally, this method requires a non-adversarial subject $x$, a sample from another target class $x_t$, and a classifier $f$ such that $f(x)\neq f(x_t)$. AMDR attempts to find an adversary $x^\prime$ that remains similar to $x$ but has an internal representation similar to $x_t$. The optimization is formulated as
\begin{equation}\label{eq:admr_attack}
    \min_{x^\prime : \|x-x^\prime\|_\infty\leq \epsilon} \|f_l(x_t)-f_l(x^\prime)\|_2,
\end{equation}
where $f_l(\cdot)$ represents the intermediary output of $f$ at the $l^\textrm{th}$ layer. As opposed to the L-BFGS attack, the ADMR attack method requires a target sample $x_t$ (rather than a target label) and a particular hidden layer $l$. This method has been shown to fool architectures such as AlexNet \citep{alexnet}, CaffeNet \citep{caffe}, and VGG variants \citep{vggvariants} on the ImageNet \citep{imagenet, imagenet2} and Places205 \citep{places} datasets. 

Other notable white box optimization-based attack methods in this category include DeepFool \citep{deepfool} and the Carlini-Wagner \citep{Carlini2017TowardsET} attacks. The former is based on finding the closest (nonlinear) decision boundary to a considered sample $x$ for a multi-class classifier. The minimal perturbation is then chosen to push the sample over the closest decision boundary. \citet{Carlini2017TowardsET} proposes a family of attacks that uses a similar notion of finding the smallest perturbation to a sample that yields the desired target classification. The authors develop different reformulations of the constraint that requires the perturbed sample $x^\prime$ to result in a model prediction of class $t$. 

\subsubsection{White-Box Gradient-Based Attacks.}\label{sss:whiteboxgradient} The family of white-box gradient-based attack methods operate on the notion of finding a perturbation direction that increases the training loss function of the target model. While choosing a perturbation as such doesn't guarantee misclassification, the target model's classification confidence will decrease. 

The fast gradient sign method (FGSM) \citep{goodfellow2015explaining} is an early proponent of this idea. It computes the gradient of the target model's loss function, and perturbs the instance $x$ according to the signs of the elements of the gradient:
\begin{equation}\label{eq:fgsm}
    x^\prime = x +\epsilon \cdot \textrm{sign}\left(\nabla_x \mathcal{L}(x, y)\right).
\end{equation}
Here $\epsilon$ is a small constant scalar that determines the size of perturbation. Unlike white-box optimization-based methods where difficult optimization problems have to be solved to generate adversarial examples, in FGSM the main computational complexity lies in evaluating the gradient of the loss function. When considering deep networks, this can be efficiently computed using the back-propagation algorithm \citep{Rumelhart1986LearningRB}. Moreover, since FGSM relies upon computing the gradient with respect to the model's loss, it can be applied to fool any machine learning method. Variants of FGSM include fast gradient value method \citep{fgvm}, basic iterative method (BIM) \citep{bim}, momentum iterative FGSM \citep{boostfgsm}, and R+FGSM \citep{rfgsm}. In \citep{Huang2017AdversarialAO}, the authors demonstrate how adversarial examples generated by FGSM can be used to fool neural network parameterized policies in reinforcement learning. FGSM and its variants are single-step methods that assume linearity. Alternatively, the projected gradient descent (PGD) attack method \citep{madry2018towards} better explores the nonlinear landscape and thereby gives an improved chance of finding the worst-case adversarial example. PGD uses several iterations to find the perturbation that maximizes the loss of the model for a particular sample while keeping the perturbation small. In each iteration, this method takes a gradient step in the direction of the greatest loss and subsequently projects back onto an $\ell_p$-norm ball that constrains the perturbation.  

\subsubsection{Black-Box Attacks.}\label{sss:blackbox}
So far we have discussed white-box attacks that leverage knowledge of the inner workings of the target model. Black-box methods assume that attackers can only query the target model.

The substitute black-box attack (SBA) \citep{sba} is one of the first proposed methods of this ilk. SBA generates a synthetic dataset labeled by the black-box model to then learn a substitute model as a means to imitate the target. Once the substitute model is trained, any of the white-box attacks discussed above can be used to generate adversarial examples. This method leverages the cross-model transferability property of adversarial examples. Boundary Attack \citep{Brendel2018DecisionBasedAA} is another black-box method used to find an adversarial example reminiscent of the sample $x$. In this method, an input $x^\prime$ sampled from another class is perturbed towards and along the decision boundary between the class of $x$ and its adjacent classes until the perceptual difference between $x$ and $x^\prime$ is sufficiently minimized.
 \citet{zoo} proposes zeroth order optimization (ZOO) attacks to directly estimate the gradients of the targeted deep network model in order to  generate adversarial examples.
In one-pixel attacks \citep{opa}, the target model is fooled by only changing a limited number of pixels of an image sample.

\subsection{Adversarial Training Methods}
So far, we have discussed a variety of methods developed to generate adversarial examples that have been shown to fool vulnerable target models. Adversarial training addresses this issue by making models robust towards worst-case adversarial examples. Therefore, the basic idea of adversarial training is to account for such worst-case perturbations by incorporating these during model training.
\subsubsection{FGSM Adversarial Training.}
FGSM Adversarial training (FGSM-AT) \citep{goodfellow2015explaining} is one of the earliest adverserial training methods, which guards against adversarial examples by augmenting these to the training dataset during each training iteration. While in FGSM-AT the adversarial examples are generated using FGSM \citep{goodfellow2015explaining}, in practice this defense methodology can also be extended to other gradient-based attack methods such as BIM \citep{kurakin2017adversarial}. The training objective for such adversarial training is of the form
\begin{equation}\label{eq:fgsm-at}
    \alpha\mathcal{L}(x, y) + (1-\alpha)\mathcal{L}(x^\prime,y),
\end{equation}
where $\alpha$ is a weighting parameter that trades off the importance of minimizing the loss with respect to the true sample $x$ and the adversarial sample $x^\prime$ generated by the attack method of choice. In practice $\alpha$ is often set to $0.5$ \citep{Wiyatno2019AdversarialEI}. While FGSM-AT has demonstrated effectiveness in defending against single-step attacks on large datasets \citep{kurakin2017adversarial}, models trained with this method have still shown to be vulnerable to multi-step attacks such as BIM or PGD. \citet{kurakin2017adversarial} found that models trained on single-step attacks such as FGSM can suffer from the issue of \textit{label leakage}. Here, it has been shown that the model trained on such single-step adversaries performs better on such adversaries than on clean data during evaluation. This suggests that examples generated by single-step adversarial attack methods may be too simple as the considered models are able to overfit on these \citep{Wiyatno2019AdversarialEI}.

\subsubsection{PGD Adversarial Training.}
Projected gradient descent adversarial training (PGD-AT) was proposed by \citet{madry2018towards} as a variant to adversarial training methods such as FGSM-AT \citep{goodfellow2015explaining}. PGD-AT is based on training against ``worst-case'' adversaries. The training objective is formulated as
\begin{equation}\label{eq:pgd-at}
    \min_\theta \max_{x^\prime : \|x^\prime - x\|\leq \epsilon} \mathcal{L}(x^\prime, y),
\end{equation}
where we select model parameters $\theta$ to minimize the worst-case perturbation within a norm-ball of radius $\epsilon$ around sample $x$. This minimax problem is aligned with the robust optimization framework where the aim is to minimize the impact of the adversary's worst-case moves. Note that PGD-AT is equivalent to setting $\alpha = 0$ in objective \eqref{eq:fgsm-at} with $x^\prime$ being the worst-case adversarial example.
\citet{madry2018towards} show that adversaries generated using the PGD attack are indeed the worst-case adversarial examples. They demonstrate that even after restarting PGD from random starting points, the model's loss plateaus to similar levels. The reason behind this is suggested to be numerous local maxima which are in turn quite similar to the global maxima \citep{madry2018towards}. Therefore, the authors argue that training against PGD adversarial examples approximately solves the minimax problem \eqref{eq:pgd-at}.
While PGD-AT has been shown to be particularly effective for large scale models \citep{madry2018towards}, some drawbacks still exist. For one, it has been shown that PGD-AT significantly compromises the model's accuracy on clean samples \citep{tsipras2018robustness}. Moreover, since the PGD attack is a multi-step method with a costly projection operation at each iteration, this method is subject to a higher computational cost compared to single-step adverarial training methods.

\subsubsection{Ensemble Adversarial Training.}
Another variant of adversarial training, proposed by \citet{tramer2018ensemble}, is known as ensemble adversarial training. Here, a model is retrained using adversarial examples used to target other pre-trained models. It is suggested that the separation of generating the adversarial examples and the actual target model mitigates the overfitting issue seen in single-step adversarial training methods such as FGSM-AT. As the process of generating adversarial examples is independent of the model that is being trained, it is argued that ensemble adversarial training is more robust to black-box attacks compared to white-box adversarial training.
\subsubsection{Convex Adversarial Training.}
A more recent form of adversarial training is based on leveraging convex reformulations of the two-layer ReLU network training problem which are shown to have equivalent optimal values to the original non-convex problem \citep{Pilanci20a, Pilanci20b, Ergen21b}. In \citep{Bai2022PracticalCF,egotwolayernetworks}, the authors develop convex robust optimization problems for the adversarial training of two-layer ReLU networks, with a particular focus on the cases of hinge loss (for binary classification) and squared loss (for regression).
\section{Robustness Certification}
\label{sec: certification}
Over recent years, we have seen an ``arms race'' develop in the robustness literature; for every proposed defense, it seems as though a new attack is developed that is strong enough to defeat the defense \citep{carlini2017adversarial,kurakin2017adversarial,athalye2018obfuscated,uesato2018adversarial,madry2018towards}. This has led researchers to consider \emph{certified robustness}, which is a mathematical proof that the model yields reliable outputs when subject to a certain threat model on the input. This threat model has been considered in both a stochastic framework, where the input uncertainty is random, as in \citet{mangal2019robustness,weng2019proven,webb2018statistical,fazlyab2019probabilistic,anderson2022data}, and an adversarial framework, where the input uncertainty is maliciously designed to fool the model. In this survey, we focus on the adversarial case, as it remains the standard threat model considered in the literature.

Certified defenses can roughly be categorized into methods based on satisfiability modulo theories (SMT), optimization- and relaxation-based bounds, and randomized smoothing. SMT-based methods such as \citet{katz2017reluplex,amir2021smt} comprise some of the earliest robustness certificates, but, as they are combinatorial in nature, the field has turned towards more tractable bounding techniques. These tractable certificates are the ones we focus on going forward.

\subsection{Optimization-Based Methods}

Consider a general neural network $f\colon \mathbf{R}^d \to \mathbf{R}^n$. The threat model considered in optimization-based methods is one in which the input to the neural network is contained in some \emph{input uncertainty set} $\mathcal{X}\subseteq \mathbf{R}^d$, but is otherwise unknown. It is common in the literature to assume that $\mathcal{X}$ is either a compact convex polytope or that $\mathcal{X} = \{x\in\mathbf{R}^d : \| x - \overline{x} \|_p \le r\}$ for some nominal input $\overline{x}\in\mathbf{R}^d$ and some attack radius $r > 0$. The goal of an adversary is to find an input $x^\star \in \mathcal{X}$ to achieve the worst-case output $f(x^\star)$, e.g., to cause misclassification in a classification task. On the other hand, the goal of the defender is to certify that all outputs $f(x)$ corresponding to inputs $x\in\mathcal{X}$ are ``safe'' in some sense. Safety can generally be modeled by a subset $\mathcal{S}$ of the output space $\mathbf{R}^n$, which is commonly taken to be a (possibly unbounded) polytope. For example, suppose that $f$ is an $n$-class classifier and that $f(x)_i$ represents the confidence associated to class $i$, so that the class assigned to an input $x$ is given by $\argmax_{i\in\{1,\dots,n\}} f(x)_i$. Then, if $\overline{x}$ is classified into class $i^*$ and we seek to certify that adversarial perturbations of $\overline{x}$ are not misclassified, then the problem amounts to showing that $f(\mathcal{X})\subseteq \mathcal{S}$ with the safe set being the half-space $\mathcal{S} = \{y\in\mathbf{R}^n : c^\top y \ge 0\}$ with $c = e_{i^*} - e_i$. We will assume that the safe set is a half-space going forward.

\begin{figure}[tbh]
    \centering
    \includegraphics[width=0.6\linewidth]{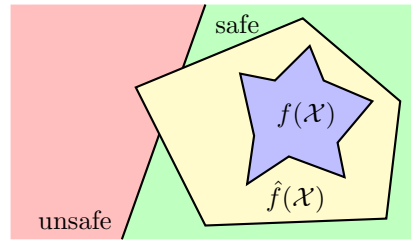}
    \caption{Certifying the safety of a convex outer-approximation $\hat{f}(\mathcal{X})$ of the nonconvex output set $f(\mathcal{X})$ is sufficient to prove robustness. If the convex approximation is too loose, it may enter the unsafe region and fail to certify \citep{anderson2021partition}}
    \label{fig: convex_relaxations}
\end{figure}

With these preliminaries in place, we may now formally define the robustness certification problem as the following optimization problem:
\begin{equation}
    p^\star = \inf_{x\in\mathcal{X}} c^\top f(x).
    \label{eq: optimization_certificate}
\end{equation}
If the optimal value $p^\star$ for \eqref{eq: optimization_certificate} is nonnegative, then $c^\top f(x) \ge 0$ for all $x\in\mathcal{X}$, implying that $f(\mathcal{X})\subseteq \mathcal{S}$, so all inputs are certified to yield safe outputs. Notice though, that most neural networks $f$ are nonconvex functions, making \eqref{eq: optimization_certificate} a nonconvex optimization problem even for convex $\mathcal{X}$. In fact, the certification problem \eqref{eq: optimization_certificate} for neural networks is proven to be NP-hard \citep{katz2017reluplex,weng2018towards}. Thus, most works make the certification procedure more tractable by either developing convex relaxations of \eqref{eq: optimization_certificate}, or directly bounding and analyzing $f$ over the set $\mathcal{X}$. Notice that, if $\hat{f}(\mathcal{X}) \supseteq f(\mathcal{X})$ and $p' = \inf_{y\in \hat{f}(\mathcal{X})} c^\top y$ is a convex relaxation of $p^\star$, then $p' \le p^\star$, so to prove the robustness of $f$ with respect to the threat model $\mathcal{X}$ it suffices to show that $p' \ge 0$. This approach to certification is illustrated graphically in Figure \ref{fig: convex_relaxations}. We now discuss such methods in more detail.

\subsubsection{Linear Programming.}
In the context of $\relu$ neural networks, \citet{wong2018provable} proposes to relax the nonlinear $\relu$ equality constraints in \eqref{eq: optimization_certificate} to their convex upper envelop. This is accomplished by introducing three linear inequalities associated to every neuron in the network. The approach is simple, scalable, and capable of being used to learn provably robust networks, but in general the feasible set grows crudely with the depth of the network, turning researchers to consider more sophisticated relaxations. Subsequent works consider fast linear bounds on $\relu$ activations \citep{weng2018towards}, generalizing the linear bounding technique past $\relu$ activations \citep{zhang2018efficient,dvijotham2018dual,wong2018scaling}, and combining linear bounds with symbolic interval analysis to attain tighter relaxations \citep{wang2018efficient}. \citet{salman2019convex} goes on to unify these linear programming-based methods and show that they suffer from a so-called ``convex relaxation barrier'' preventing tight certification under this framework. The paper \citet{tjandraatmadja2020convex} proposed to overcome this barrier by using a tightened, joint linear relaxation of $\relu$ networks instead of the prior neuron-wise relaxations.

\citet{gowal2018effectiveness} considers using interval bounds on each layer's activation values to obtain a bound on the optimal value of \eqref{eq: optimization_certificate}, and incorporate this efficient procedure into the training process to learn certifiably robust networks. Other works such as \citet{hein2017formal,ruan2018reachability,fazlyab2019efficient} derive bounds and estimates on the Lipschitz constant of a neural network, which is then able to be used to assess the network's robustness.

\subsubsection{Semidefinite Programming.}

Since the feasible set of linear programming relaxations grows crudely as the size of the network grows, researchers have considered nonlinear convex relaxations in order to tighten the certification. The authors of \citet{raghunathan2018semidefinite} reformulate \eqref{eq: optimization_certificate} for $\relu$ networks as an optimization over a lifted positive semidefinite matrix variable with a rank-1 constraint. By dropping the rank constraint, they arrive at a semidefinite program, which is found to outperform linear programming relaxations---especially for deep networks---increasing the percentage of certified MNIST examples at an $\ell_\infty$-attack radius of $r=0.1$ by over 80\% for networks learned via PGD adversarial training. The work \citet{zhang2020tightness} theoretically showed that this semidefinite relaxation is tight for a single hidden layer under mild technical assumptions. In \citet{fazlyab2020safety}, quadratic constraints are used to describe a neural network's activation function, and the S-procedure is then used to arrive at a tightened semidefinite relaxation. Although these semidefinite relaxations yield good bounds on robustness, they are not generally applicable to large-scale settings.

\subsubsection{Partitioned and Mixed-Integer Programming.}

In order to reach a middle ground attaining both the computational tractability of simple linear programming relaxations and the tightness of more sophisticated relaxations, a line of works has been developed that make use of partitioning schemes and mixed-integer optimization formulations. For example, \citet{tjeng2017evaluating} reformulates the certification problem \eqref{eq: optimization_certificate} for $\relu$ networks into a mixed-integer linear program that scales up to networks of $100,000$ neurons, and is able to solve for $p^\star$ exactly. The paper \citet{bunel2018unified} builds on this formulation and proposes an efficient branch-and-bound algorithm for estimating the optimal value.

Related to the mixed-integer formulation is the partitioned programming approach, wherein the input set $\mathcal{X}$ is partitioned into smaller subsets, and the optimization \eqref{eq: optimization_certificate} is solved over each subset individually to arrive at an overall global bound on $p^\star$. The partitioning is effective since convex relaxations tend to become tighter as the feasible set of the optimization becomes smaller. An example of this approach is \citet{xiang2018reachability}, where the input uncertainty set is partitioned into hyperrectangles, leading to tightened certification of networks with general activation functions. The work \citet{rubies2019fast} uses Lagrangian duality to motivate a partition of box-shaped input uncertainty sets along coordinate axes. In a similar vein, \citet{everett2020robustness} theoretically characterizes the reduction in volume of the resulting outer-approximation of $f(\mathcal{X})$ upon using axis-aligned partitioning in the input space. In \citet{anderson2020tightened}, the authors upper-bound the worst-case relaxation error induced by the linear programming approach of \citet{wong2018provable}, and then derive a closed-form expression for the optimal partition of $\mathcal{X}$ that minimizes this worst-case error. \citet{ma2021sequential} propose a partitioning scheme that sequentially reduces the relaxation error of the semidefinite programming approach and characterizes its effect on the feasible set geometry. The work \citet{wang2021beta} develops a parallelizable bound propagation method that optimizes over partitioning parameters leading to state-of-the-art certification results, outperforming semidefinite relaxations on MNIST and CIFAR-10 benchmarks in terms of both time and certified accuracy, and winning the 2021 Verification of Neural Networks Competition \citep{bak2021second}.

\subsection{Randomized Smoothing}

\begin{figure}[tbh]
    \centering
    \includegraphics[width=0.6\linewidth]{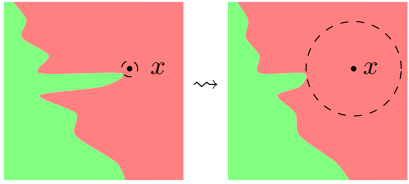}
    \caption{Randomized smoothing smooths out jagged regions of the base classifier decision boundary (left) to certifiably increase the robust radius around inputs (right) \citep{anderson2022towards}}
    \label{fig: randomized_smoothing}
\end{figure}

Randomized smoothing, popularized in \citet{lecuyer2019certified,li2019certified,cohen2019certified}, is a probabilistic method to robustify a classifier $f\colon\mathbf{R}^d\to\{1,2,\dots,n\}$ with deterministic robustness guarantees. Formally, we consider a hard-classifier $f(x) \in \argmax_{i\in\{1,2,\dots,n\}} g_i(x)$ for some associated soft-classifier $g\colon\mathbf{R}^d\to\mathbf{R}^n$. The majority of works on randomized smoothing assume that the image of $g$ is the probability simplex in $[0,1]^n$---a realistic assumption that simplifies the analyses. Now, letting $\mu$ denote some probability distribution on the input space $\mathbf{R}^d$, randomized smoothing proposes to replace the classifier $f$ by the following ``smoothed classifier'':
\begin{equation*}
    f^\mu(x) \in \argmax_{i\in\{1,2,\dots,n\}} g_i^\mu(x), \quad
    g_i^\mu(x) = \mathbf{E}_{\epsilon\sim\mu} g(x+\epsilon).
\end{equation*}
Intuitively, randomized smoothing intentionally corrupts an input $x$ by additive random noise, feeds that through the base classifier, and uses the average prediction of these noisy inputs as the actual prediction for $x$. The idea behind this process is to ``average out'' any potential dangerous, yet unlikely adversarial perturbations, in effect robustifying the classification scheme. This is graphically illustrated in Figure \ref{fig: randomized_smoothing}.

The most common distribution $\mu$ to use is an unbiased symmetric Gaussian, i.e., $N(0,\sigma^2 I_d)$ for some user-prescribed variance $\sigma^2$. We first describe this setup, and then move on to discuss variants and on-going work to generalize the smoothing framework.

\subsubsection{Symmetric Gaussian Smoothing.}

Symmetric Gaussian smoothing, where $\mu = N(0,\sigma^2 I_d)$, is the most common form of randomized smoothing. This is the method introduced in \citet{lecuyer2019certified,li2019certified}, and popularized in \citet{cohen2019certified} with provably tight robustness guarantee, which we recall in the following theorem.

	\begin{theorem}[\citealp{cohen2019certified,zhai2020macer}]
		\label{thm: cohen}
		Let $\sigma^2$ be a positive real number and consider $\mu = N(0,\sigma^2 I_d)$. Consider a point $x\in\mathbf{R}^d$ and let $y = f^\mu(x)\in\argmax_{i\in\{1,\dots,n\}} g^\mu_i(x)$ and $y' \in \argmax_{i\in\{1,\dots,n\}\setminus\{y\}} g^\mu_i(x)$. Then $f^\mu(x+\delta) = y$ for all $\delta\in\mathbf{R}^d$ such that
		\begin{equation*}
			\|\delta\|_2 \le r^\mu(x) \coloneqq \frac{\sigma}{2} \left( \Phi^{-1}(g^\mu_y(x)) - \Phi^{-1}(g^\mu_{y'}(x)) \right).
		\end{equation*}
	\end{theorem}

    Theorem \ref{thm: cohen} says that any attack of $\ell_2$-norm less than $r^\mu(x)$ cannot change the prediction of the nominal input $x$. This certified robustness guarantee enables the use of smoothed models in safety-critical systems with confidence in their reliability, e.g., policies in reinforcement learning or neural network-based controllers \citep{wu2021crop,kumar2021policy}. Due to its efficient statistical nature, randomized smoothing is able to scale to extremely large settings, even allowing for the certification of ImageNet models \citep{cohen2019certified}, which had previously been intractable for non-smoothing-based certification methods.
    
    Follow-up works attempt to increase the size of the certified region using a variety of different approaches. For example, \citet{salman2019provably} incorporates adversarial training into the Gaussian smoothing scheme to obtain state-of-the-art $\ell_2$-norm certified radii. On the other hand, \citet{zhai2020macer} develops a tractable method to maximize the certified radius over the smoothing variance $\sigma^2$. The authors of \citet{zhang2020black} look at optimizing the base classifier in order to achieve maximal certified radii, and their experiments show a notable increase in certified robust accuracy over the baseline \citet{cohen2019certified}. More recently, the authors of \citet{pfrommer2022projected} incorporated dimensionality reduction of the neural network's input in order to perform Gaussian smoothing in a lower-dimensional space. This in effect projects-out perturbations that are normal to the natural data manifold, enlarging the certified region along such directions in the input space.
    
    There has also been a line of works demonstrating negative results for randomized smoothing. For example, the works \citet{tsipras2018robustness,krishnan2020lipschitz,yang2020closer,gao2020analyzing} all point to an inherent accuracy-robustness tradeoff for smoothed models. That is, as $\sigma^2$ increases, the model becomes smoother, and hence its regions of constant classification become larger, but accuracy on clean test samples ends up suffering once $\sigma^2$ becomes too large. Another intriguing negative result is the ``shrinking phenomenon'' discovered in \citet{mohapatra2021hidden}, where, as $\sigma^2$ increases, bounded decision regions and those contained within a narrow cone tend to shrink, causing imbalances in classwise accuracies.

\subsubsection{Alternative Smoothing Distributions.}

As symmetric Gaussian smoothing only yields certified regions taking the form of an isotropic $\ell_2$-ball, researchers have worked on generalizing randomized smoothing to certify alternatively shaped regions. The work \citet{tengl1} uses a Laplacian smoothing distribution to certify $\ell_1$-norm balls, \citet{levine2020wasserstein} certifies against attacks defined in terms of the Wasserstein distance, \citet{lee2019tight} uses discrete distributions to certify against cardinality-constrained ``$\ell_0$''-attacks, and \citet{yang2020randomized} characterize the optimal smoothing distributions corresponding to $\ell_1$-, $\ell_2$-, and $\ell_\infty$-attacks. The authors of \citet{erdemir2021adversarial} extend randomized smoothing to an asymmetric variant in order to certify anisotropic regions of the input space.

\subsubsection{Input-Dependent Smoothing.}

Since inputs close to the base classifier decision boundary may require small smoothing variance to maintain clean accuracy, yet inputs far from the decision boundary may permit large variance so as to increase their corresponding certified radii, a recent push to allow for input-dependent smoothing distributions has been made. In particular, the smoothing distribution $\mu$ is replaced by $\mu(x)$, which is a probability distribution that may differ between inputs $x$.

In order to derive mathematically valid robustness certificates, additional assumptions on the map $x\mapsto \mu(x)$ must be made. For example, the works \citet{alfarra2020data,wang2021pretrain} choose $\mu(x)$ to be a normal distribution such that the variance is optimized for particular data points locally in the vicinity of $x$. However, this approach amounts to a classifier that is dependent on the order of incoming inputs for which the variances are optimized, in effect introducing another uncertainty into the classification scheme as well as a vulnerability that an adversary may exploit. The work \citet{eiras2021ancer} generalizes this approach to certify anisotropic ellipsoids, choosing the Gaussian covariance to maximize the volume of the ellipsoids in a pointwise fashion.

In \citet{sukenik2021intriguing}, it is shown that, in general, input-dependent randomized smoothing suffers from the curse of dimensionality. The authors propose a parameterization of the smoothing scheme that avoids the order-dependence of the classifier from the prior works, but without notable improvement in the certified accuracies over conventional input-independent randomized smoothing. The authors of \citet{anderson2022certified} argue that effective randomized smoothing schemes should be designed with the underlying data distribution in mind, and they show that in order to achieve this the smoothing distribution must necessarily be input-dependent and have a nonzero mean. Accordingly, they develop an input-dependent smoothing scheme for binary classifiers that is optimal up to a first-order approximation of the base classifier. Along similar lines, \citet{anderson2022towards} formulates the optimal randomized smoothing problem as the optimization over general (not necessarily Gaussian) distribution-valued maps $x\mapsto \mu(x)$ subject to constraints on their Lipschitz continuity. The Lipschitz condition is utilized in order to derive mathematically valid certified radii for the resulting model. They show that the optimal distribution can be approximated by a semi-infinite linear program, and prove that the resulting problem satisfies nontrivial strong duality, allowing for more tractable computations. Although there has been much research developed surrounding randomized smoothing and its variants, the problem of \emph{exactly} optimizing the smoothing distribution in a manner scalable to large datasets such as ImageNet appears to remain open.
\section{Conclusions and Directions for Future Research}
\label{sec: conclude}
While there have been great strides in recent years on finding new methods to generate and defend against adversarial attacks, the fundamental question regarding why deep architectures are vulnerable to such adversarial examples still remains open. A unified theory explaining the vulnerability as well as properties such as cross-model transferability will in turn provide valuable insight into effective training methods that are robust to a variety of attacks. Another direction of future research relates to standardizing robustness evaluation across different attacks. Currently, defense approaches claim that they can guard against certain attacks while failing against others. A standardized evaluation framework will enable a fair judgement on the defense efficacy of robust training methods for different attacks.

In the realm of certification, it appears to remain an open problem to derive optimization-based methods for general (non-$\relu$) activations that are tight enough to yield meaningful guarantees while being computationally tractable enough to scale to large benchmarks like ImageNet. The problem of optimizing input-dependent randomized smoothing schemes is a promising direction for certifiably robustifying classifiers in the large-scale setting, but more work must be done in order to tractably solve for such distributions without resorting to heuristics such as locally-constant input-to-distribution maps.


\bibliography{ifacconf} 


\end{document}